\documentclass[10pt,twocolumn,letterpaper]{article}

\usepackage[pagenumbers]{cvpr} 

\usepackage{graphicx}
\usepackage{amsmath}
\usepackage{amssymb}
\usepackage{booktabs}
\usepackage{bm}
\usepackage{subcaption}
\usepackage{multirow}
\usepackage{bbding}

\usepackage[pagebackref,breaklinks,colorlinks]{hyperref}

\usepackage[capitalize]{cleveref}

\crefname{section}{Sec.}{Secs.}
\Crefname{section}{Section}{Sections}
\Crefname{table}{Table}{Tables}
\crefname{table}{Tab.}{Tabs.}

\newcommand{\tabincell}[2]{\begin{tabular}{@{}#1@{}}#2\end{tabular}}

\begin{document}

\title{IntraQ: Learning Synthetic Images with Intra-Class Heterogeneity for Zero-Shot Network Quantization}

\author{Yunshan Zhong$^{1,2}$, Mingbao Lin$^2$, Gongrui Nan$^2$, Jianzhuang Liu$^3$, Baochang Zhang$^4$,\\ Yonghong Tian$^{5,6}$, Rongrong Ji$^{1,2,6}$\thanks{Corresponding Author}\\
$^1$Institute of Artificial Intelligence, Xiamen University\\
$^2$Media Analytics and Computing Lab, Department of Artificial Intelligence, \\School of Informatics, Xiamen University, \\ $^3$Noah’s Ark Lab, Huawei Technologies, \\ $^4$Beihang University, $^5$Peking University, $^6$Peng Cheng Laboratory\\
{\tt\small \{zhongyunshan, lmbxmu, nangongrui\}@stu.xmu.edu.cn, liu.jianzhuang@huawei.com,}\\
{\tt\small bczhang@buaa.edu.cn, yhtian@pku.edu.cn, rrji@xmu.edu.cn}
}

\maketitle

\begin{abstract}
Learning to synthesize data has emerged as a promising direction in zero-shot quantization (ZSQ), which represents neural networks by low-bit integer without accessing any of the real data. In this paper, we observe an interesting phenomenon of intra-class heterogeneity in real data and show that existing methods fail to retain this property in their synthetic images, which causes a limited performance increase. To address this issue, we propose a novel zero-shot quantization method referred to as IntraQ. First, we propose a local object reinforcement that locates the target objects at different scales and positions of the synthetic images. Second, we introduce a marginal distance constraint to form class-related features distributed in a coarse area. Lastly, we devise a soft inception loss which injects a soft prior label to prevent the synthetic images from being overfitting to a fixed object. Our IntraQ is demonstrated to well retain the intra-class heterogeneity in the synthetic images and also observed to perform state-of-the-art. For example, compared to the advanced ZSQ, our IntraQ obtains 9.17\% increase of the top-1 accuracy on ImageNet when all layers of MobileNetV1 are quantized to 4-bit. Code is at \url{https://github.com/zysxmu/IntraQ}.

\end{abstract}

\section{Introduction}
\label{sec:intro}
The increasing demands in computing power and memory footprint of deep neural networks (DNNs) raise a challenging application problem on edge computing devices such as smart phones or wearable gadgets, in which the limited hardware resource fails to support the highly complex DNNs.
A variety of methods~\cite{he2019filter,whitepaper,hinton2015distilling,lin2020hrank} have been investigated to reduce the model complexity. Network quantization, which represents the floating-point parameters and activations within the networks by low-bit integers, stands out among these methods for its significant memory reduction and more efficient integer operations.

%
%

Most existing methods explore quantization-aware training (QAT) that builds a quantizer on the premise of accessing the original complete training dataset \cite{bulat2020HighCapacity,gong2019differentiable,yang2021fracbits}. In \cite{APoT,kim2021distance,zhuang2020training}, QAT is demonstrated to be comparable or even better than its floating-point counterpart since the learned weights could be adjusted to fit the quantization operations given the access to sufficient training data~\cite{BitSplitStitching}. However, the drawbacks also stem from its reliance on training data. Specifically, in many real-world cases, the original training data is sometimes prohibitive due to deteriorating privacy and security problems. For example, people may not wish their medical records to be revealed to others, and business material is not expected to be transmitted via the internet. As such, QAT is no longer applicable. Though recent studies concentrate on post-training quantization (PTQ) \cite{li2021brecq,Upordown,BitSplitStitching} which directly quantizes DNNs using a small portion of original data, for cases such as MLaas (\emph{e.g.}, Amazon AWS and Google Cloud), it may be impossible to reach any of the training data from users~\cite{zeroq}.


\begin{figure*}[!t]
  \centering
  \vspace{3mm}
  \begin{subfigure}{0.15\linewidth}
    \includegraphics[width=\linewidth]{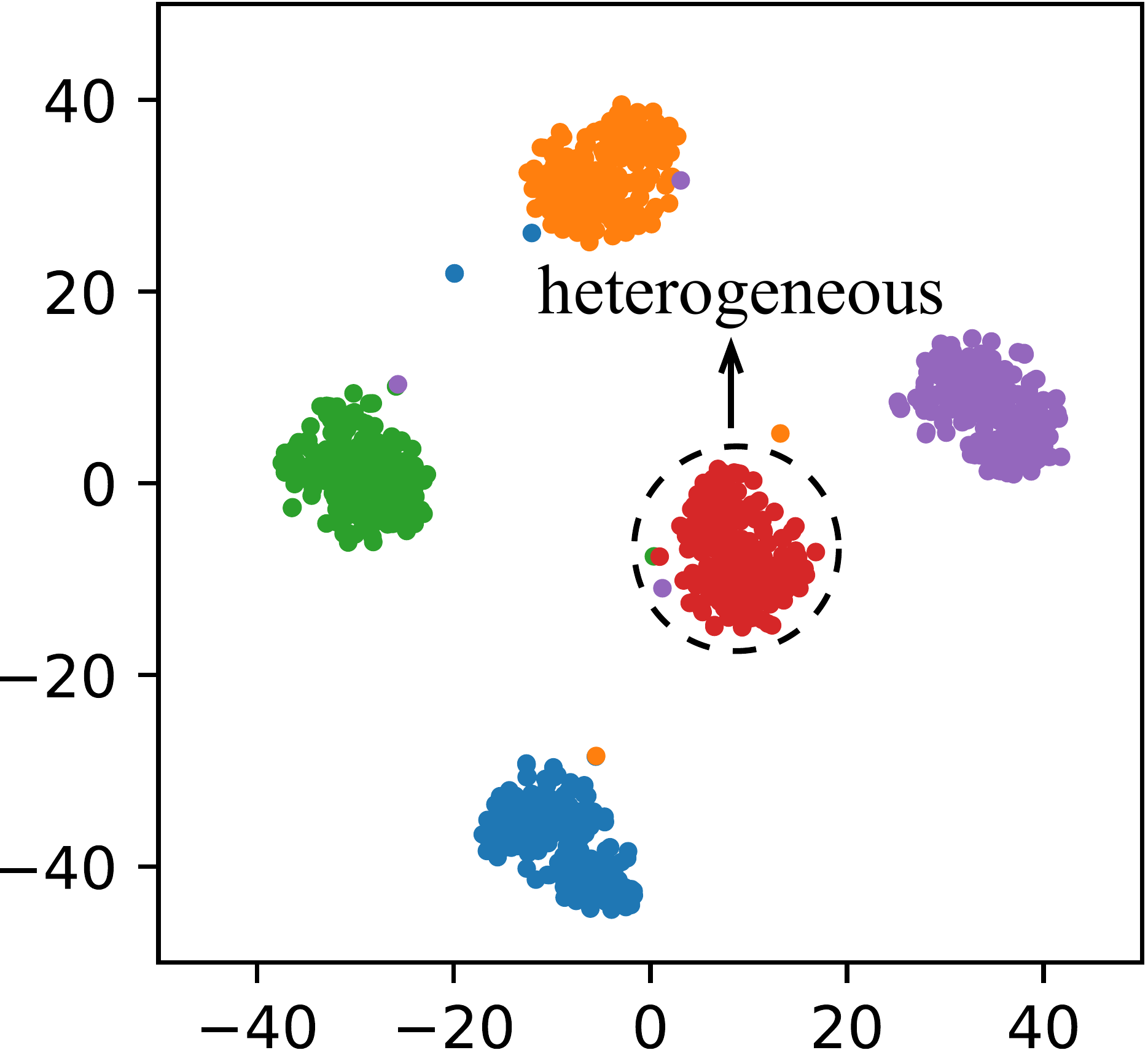}
    \caption{Real data}
    \label{phenomenon:real}
  \end{subfigure}
  \begin{subfigure}{0.15\linewidth}
    \includegraphics[width=\linewidth]{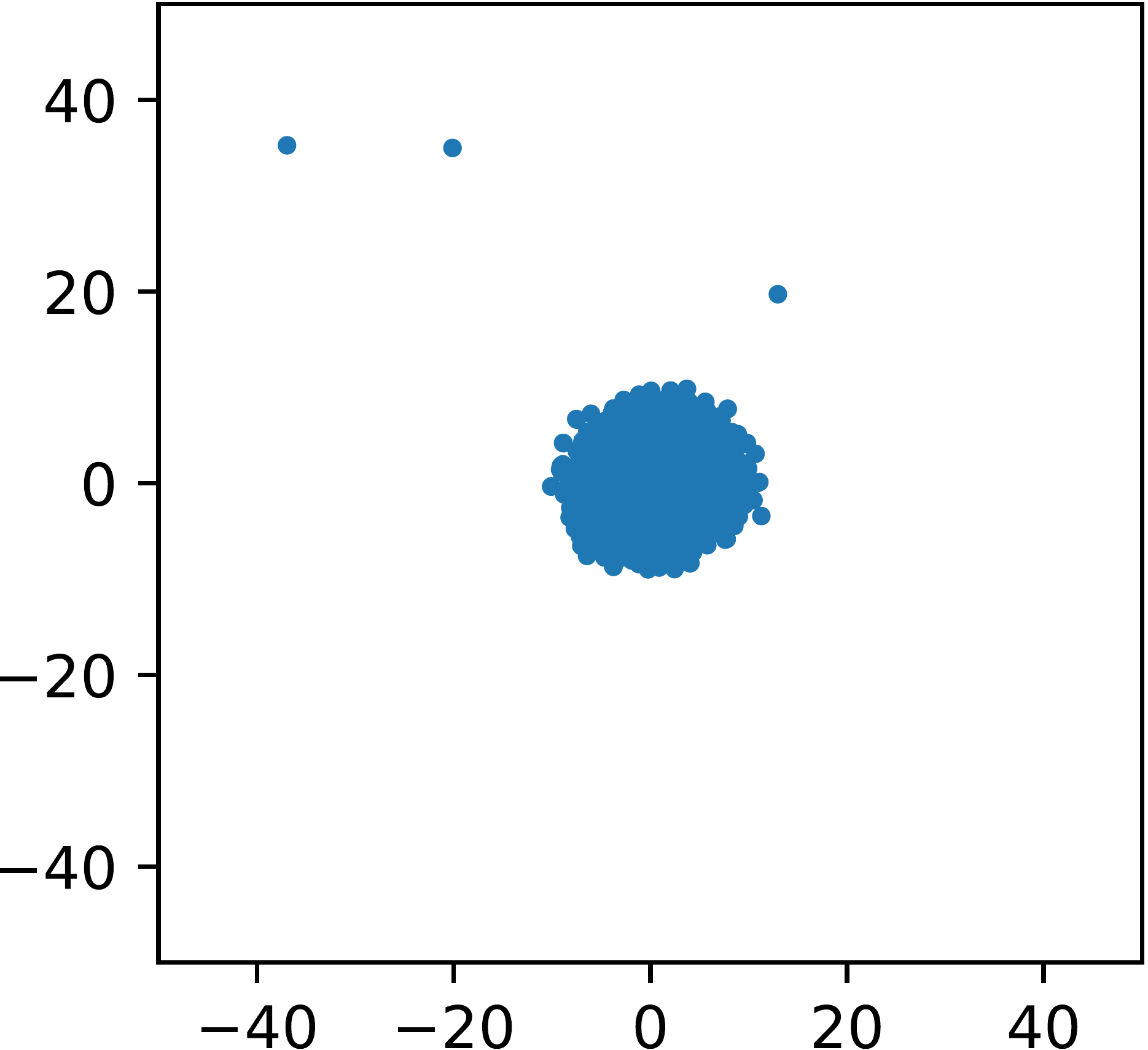}
    \caption{ZeroQ}
    \label{phenomenon:zero}
  \end{subfigure}
  \begin{subfigure}{0.15\linewidth}
    \includegraphics[width=\linewidth]{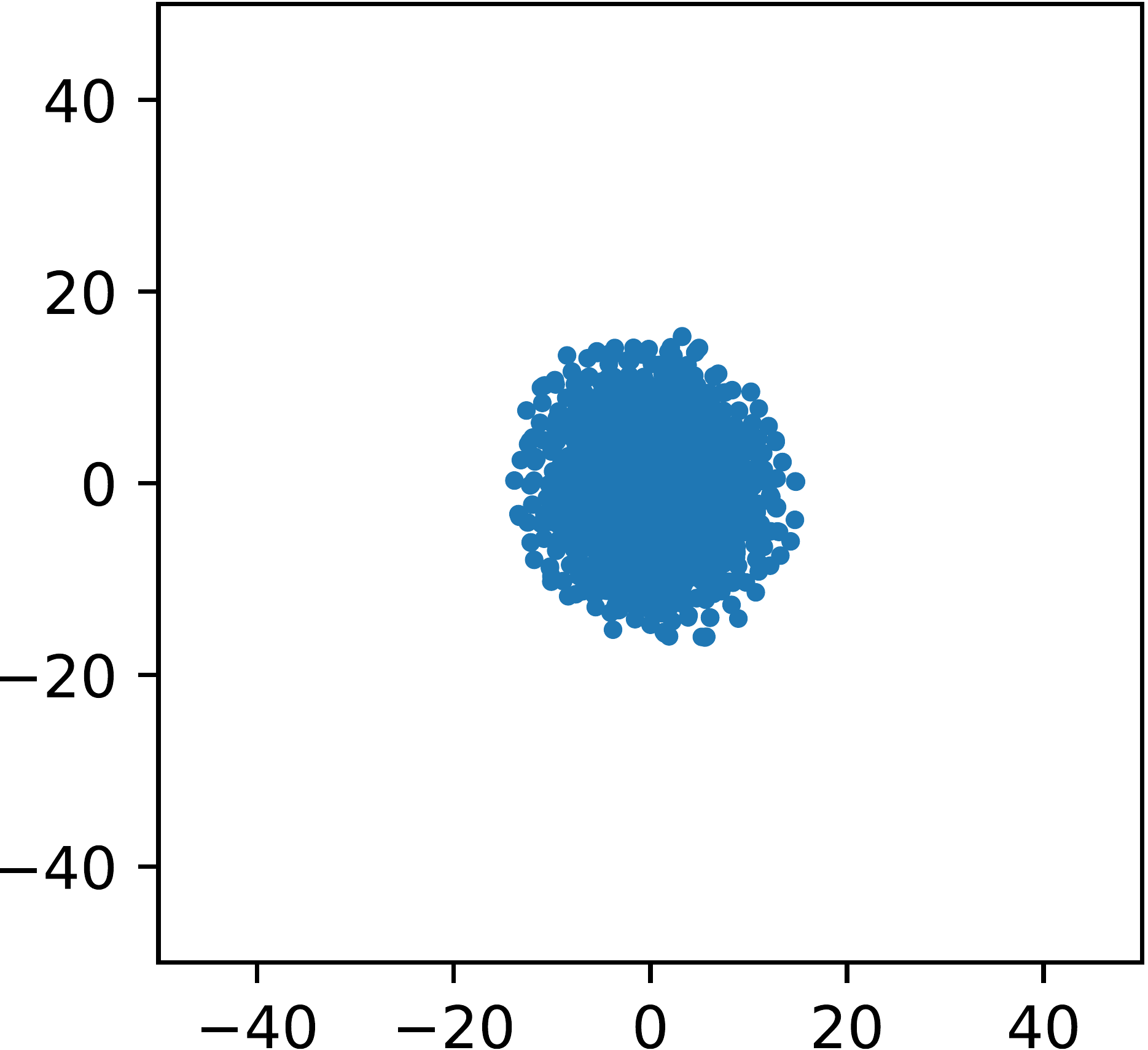}
    \caption{DSG}
    \label{phenomenon:dsg}
  \end{subfigure}
  \begin{subfigure}{0.15\linewidth}
    \includegraphics[width=\linewidth]{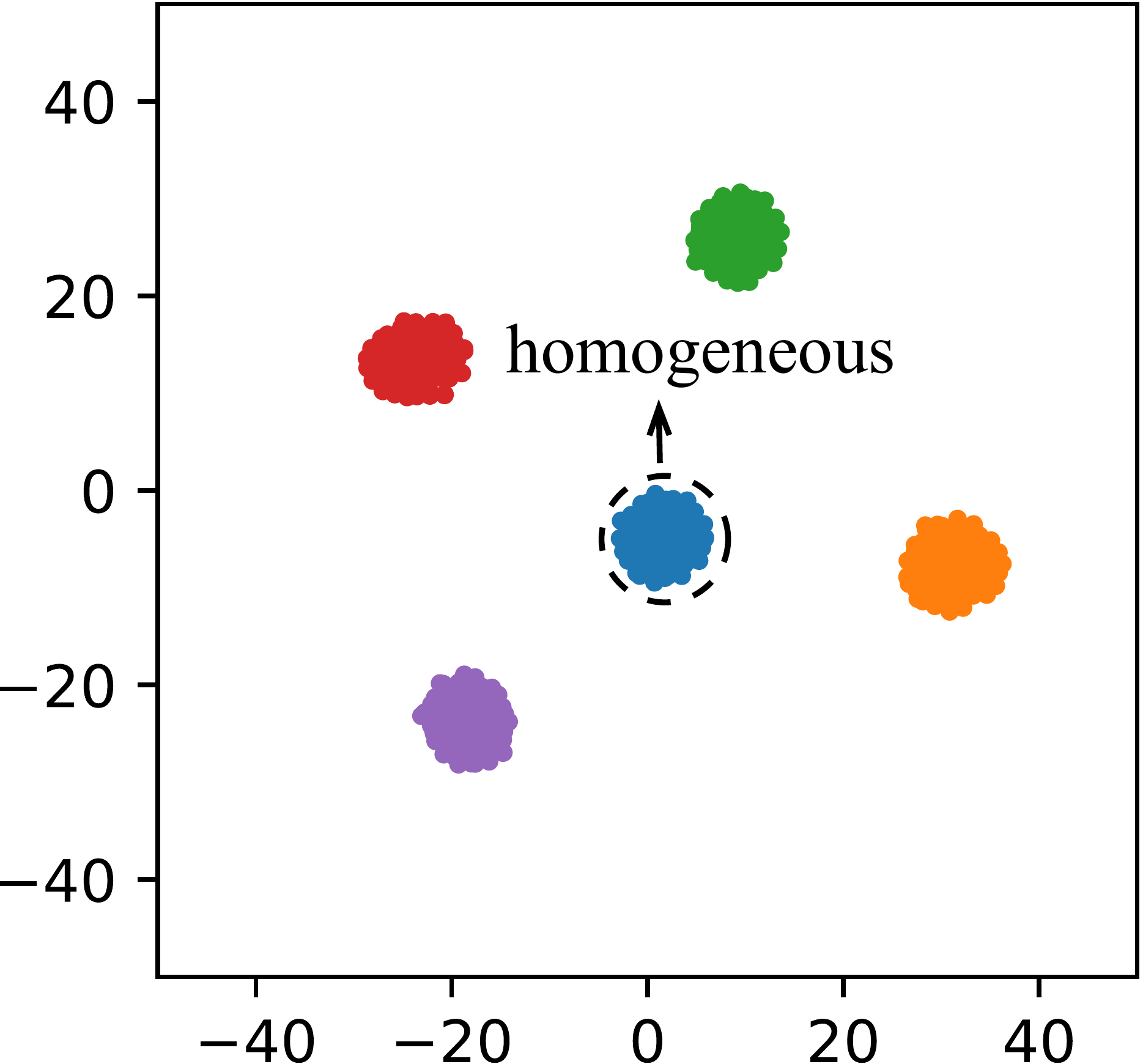}
    \caption{ZeroQ+IL}
    \label{phenomenon:zeroq+IL}
  \end{subfigure}
  \begin{subfigure}{0.15\linewidth}
    \includegraphics[width=\linewidth]{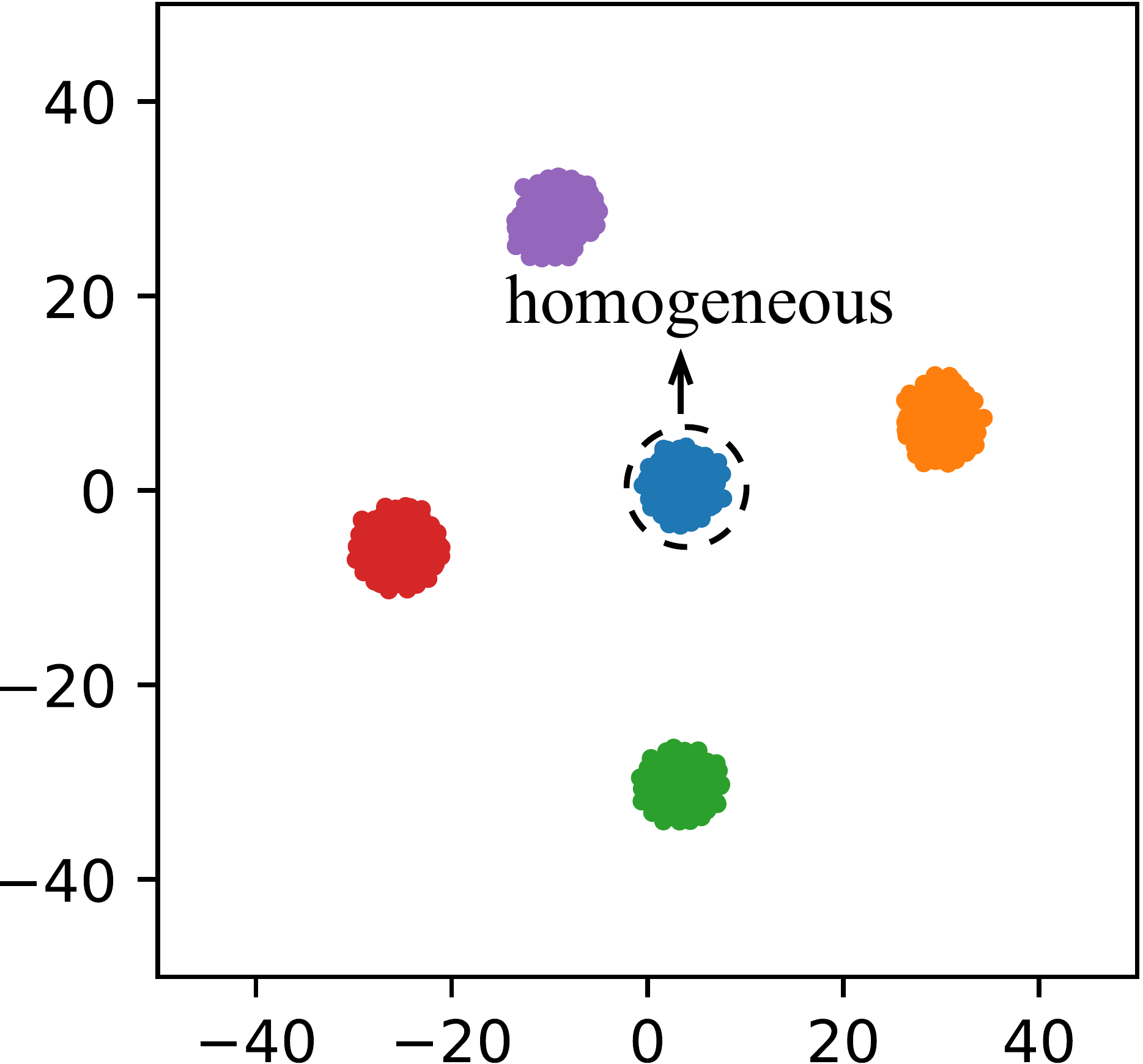}
    \caption{DSG+IL}
    \label{phenomenon:dsg+IL}
  \end{subfigure}
  \begin{subfigure}{0.15\linewidth}
    \includegraphics[width=\linewidth]{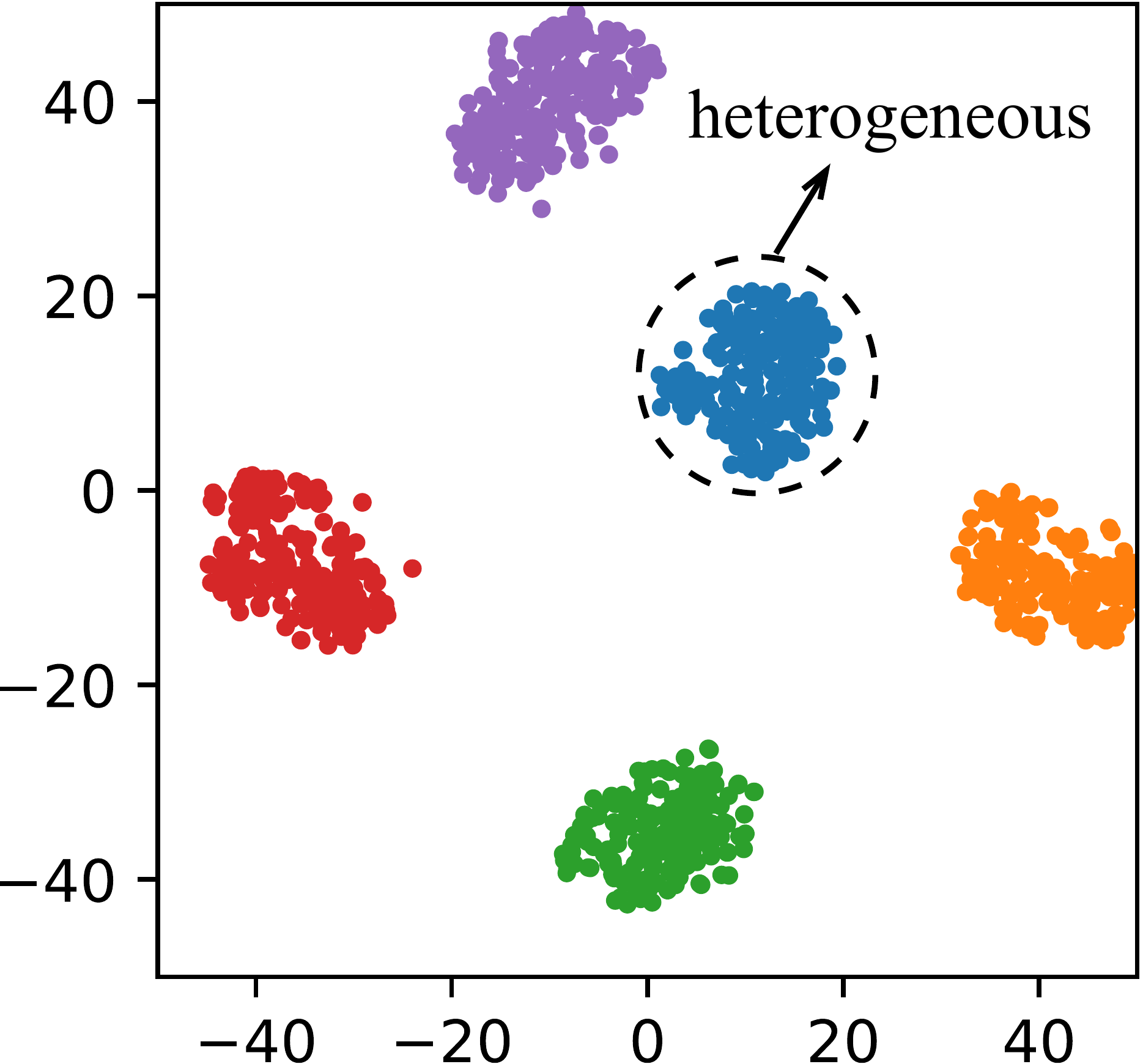}
    \caption{\textbf{IntraQ} (Ours)}
    \label{phenomenon:ours}
  \end{subfigure}
  \vspace{-0.5em}
  \caption{Feature visualization using $t$-SNE~\cite{tSNE}. We randomly sample 1,000 synthetic/real images consisting of 5 classes with 200 images per class. For ZeroQ and DSG, label information is unavailable. The features are extracted from a pre-trained ResNet-18.}
  \label{phenomenon}
  \vspace{-1.0em}
\end{figure*}

Fortunately, the research community recently has proposed zero-shot quantization (ZSQ) to quantize models without accessing real data. We empirically categorize existing studies on ZSQ into two groups. The first group concentrates on calibrating parameters without the involvement of any data. For example, DFQ~\cite{DFQ} only utilizes the shift and scale parameters $\bm{\beta}$ and $\bm{\gamma}$ stored in the batch normalization layers of the pre-trained full-precision model to compute the expected biased error on the output. Nevertheless, a simple calibration of parameters often results in severe performance degradation. This issue is even amplified for the ultra-low precision situation. For instance, only 0.10\% top-1 accuracy of DFQ on ImageNet~\cite{russakovsky2015imagenet} is reported in the appendix of \cite{GDFQ} when quantizing ResNet-18 to 4-bit.

The second group performs quantization by exploiting synthetic fake images. The involvement of fake images facilitates the training of quantized networks which are demonstrated to be superior in performance~\cite{VAEGenerate,GDFQ,DSG}.
An intuitive solution is to deploy a generator to synthesize training data~\cite{GDFQ,DQAKD,ZAQ}. However, these generator-based methods suffer a heavy overhead on computation resources since the introduced generator has to be trained from scratch for different bit-width settings.
On the contrary, many studies such as ZeroQ~\cite{zeroq} and DSG~\cite{DSG} formulate the data synthesis as an optimization problem where a random input data drawn from a standard Gaussian distribution is iteratively updated to fit the real-data distribution. The benefit of this research line is that the synthetic images can be reused to calibrate or fine-tune networks in different bit widths, resulting in a resource-friendly quantization.
Nevertheless, there still remains a non-ignorable quality gap in synthetic images when comparing the feature visualization of ZeroQ (Fig.\,\ref{phenomenon:zero}) and DSG (Fig.\,\ref{phenomenon:dsg}) with the real data (Fig.\,\ref{phenomenon:real}) since traditional Gaussian synthesis is towards fitting the whole dataset while ignoring a subtler class-wise decision boundary. Thus, the quantized models often bear large performance drops (see Sec.\,\ref{data synthesis}).

To ensure class-wise discrimination in the fake images, we apply the popular inception loss~\cite{Theknowledgewithin,yin2020dreaming} to ZeroQ and DSG, which first chooses an arbitrary label, and then performs optimization to generate label-oriented images. As a result, we observe more class-wise separable distributions of synthetic data (Fig.\,\ref{phenomenon:zeroq+IL} and Fig.\,\ref{phenomenon:dsg+IL}). This demonstrates the importance of injecting prior class information in synthetic data.
Nevertheless, we observe that the synthetic data with the inception loss does not well capture the intra-class heterogeneity. Specifically, images from the same class often contain different contents; thus features from the same class of real data scatter a lot as shown in in Fig.\,\ref{phenomenon:real}. On the contrary, those in Fig.\,\ref{phenomenon:zeroq+IL} and Fig.\,\ref{phenomenon:dsg+IL} are in a dense concentration, which indicates the synthetic images of the same class are mostly homogeneous. Consequently, the quantized model fine-tuned with these synthetic data fails to generalize well to the real-world test dataset featuring heterogeneity.

To retain the intra-class heterogeneity, in this paper, we propose a novel zero-shot quantization method, termed IntraQ. 
Motivated by the fact that the objects of interest benefiting the model learning are not always at the same scale or position in the images, we propose a local object reinforcement by randomly cropping a local patch from the synthetic image to locate the target objects, which mitigates synthesizing homogeneous images. 
Apart from heterogeneous images, we also propose to retain the intra-class heterogeneity in their feature representation. This is accomplished by introducing a marginal distance constraint to not only form class-related features but also avoid learning features concentrated on a dense area.
In contrast to the traditional inception loss with one-hot label vectors, we further devise a soft inception loss which injects a soft prior label to excavate images with more complex scenes and prevent the synthetic images from being overfitting to a fixed object.
With the above three innovative solutions, the intra-class heterogeneity is well preserved in our synthetic images as shown in Fig.\,\ref{phenomenon:ours} and significant performance improvements are observed when using only 5,120 synthetic images to fine-tune the quantized models. For instance, our IntraQ achieves 51.36\% top-1 accuracy on ImageNet when quantizing MobileNetV1 to 4-bit, leading to a increase of 9.17\% when compared with the advanced DSG~\cite{DSG} equipped with the traditional inception loss~\cite{Theknowledgewithin}.

%
%

\section{Related Work}
\label{sec:related}

In this section, we briefly review data-driven quantization and zero-shot quantization. A comprehensive discussion can be referred to the recent survey paper~\cite{gholami2021survey}.

\subsection{Data-Driven Quantization}

Both QAT and PTQ require real data to complete quantization. With abundant training images, existing QAT methods focus on designing quantizers~\cite{LSQ,APoT,jung2019learning}, training strategies~\cite{zhuang2020training,lee2021network}, dynamic quantization~\cite{jin2020adabits,shen2020fractional,yu2021AnyPrecision}, binary networks~\cite{rastegari2016xnor,qin2020forward,lin2020rotated,real2binary}, approximate gradients~\cite{gong2019differentiable,yang2019quantization}, \emph{etc}.
On the contrary, PTQ is limited to accessing a very small portion of training data~\cite{ACIQ,fang2020post,li2021brecq,Upordown,BitSplitStitching,MPwMP}. Banner~\emph{et al}.~\cite{ACIQ} combined analytical clipping, per-channel bit allocation, and bias-correction to form a 4-bit post-training method.
AdaRound~\cite{Upordown} shows that the rounding-to-nearest is not the optimal rounding function and formulates the rounding as a layer-wise quadratic unconstrained binary problem. 
In~\cite{MPwMP}, a linear combination of multiple low-bit vectors is used to approximate a full-precision weight vector. A mixed-precision network is constructed upon a quantization error-based greedy selection to adaptively decide the number of low-bit vectors.
Based on a theoretical study of the second-order loss and empirical evidence, Li~\emph{et al}.~\cite{li2021brecq} proposed a block reconstruction to regain the accuracy.

\subsection{Zero-Shot Quantization}

ZSQ accomplishes network quantization without accessing any real data. To this end, DFQ~\cite{DFQ} focuses on calibrating network parameters by utilizing the scale-equivariance property. To fix inherent bias rising from quantization~\cite{ACIQ} without data, the shift parameter $\bm{\beta}$ and scale parameter $\bm{\gamma}$ in the BN layers are used to calculate the expected biased error on the outputs. 
To boost accuracy performance, another group concentrates on synthesizing fake images. GDFQ~\cite{GDFQ} incorporates the BNS alignment loss and inception loss to train a generator for generating label-oriented images. To diversify synthetic images, DQAKD~\cite{DQAKD} trains the generator in an adversarial manner. ZAQ~\cite{ZAQ} also adversarially trains the generator with a novel two-level modeling strategy to measure the discrepancy.
In addition to the generator, the data synthesis can also be realized by optimizing the Gaussian noise. By regarding batch normalization statistics (BNS) ($\emph{i.e.}$ running mean $\bm{\mu}$ and running variance $\bm{\sigma}^2$) as the distribution indicators, ZeroQ~\cite{zeroq} optimizes the Gaussian noise such that the mean and variance of synthetic data can match the BNS in the pre-trained network. DSG~\cite{DSG} first relaxes the BNS alignment loss to prevent synthetic images from over-fitting, and then for each sample, randomly enlarges the loss term in backward propagation. Inspired by VAE~\cite{VAE}, GZNQ~\cite{VAEGenerate} regards the synthetic images as optimizable parameters and introduces ensembling to model hard samples. By approximating BNS, ~\cite{horton2020layer} estimates the fake data to determine activation ranges.

\section{Methodology}

\subsection{Preliminaries}

\subsubsection{Quantizer}

Following the settings of~\cite{GDFQ}, we use the asymmetric uniform quantizer to implement network quantization. Denoting $\bm{x}$ as weights/activations, $l$ and $u$ as the lower bound and upper bound of $\bm{x}$, we can obtain the quantized integer $\bm{q}$ as:
\begin{equation}
\bm{q} = round\big(\frac{clip(\bm{x}, l, u)}{s}\big),
\label{Equation1}
\end{equation}
where $clip(\bm{x}, l, u) = min(max(\bm{x}, l), u)$ and $round(\cdot)$ rounds its input to the nearest integer. $s = \frac{u - l}{2^b-1}$ is the scaling factor that projects a floating-point number to a fixed-point integer and $b$ is the bit-width. The corresponding de-quantized value $\bar{\bm{x}}$ can be calculated as:
\begin{equation}
\bar{\bm{x}} = \bm{q} \cdot s.
\end{equation}

For activations and weights, we use layer-wise quantizer and channel-wise quantizer respectively.

\subsubsection{Data Synthesis}\label{data synthesis}
ZSQ receives popularity mostly due to its evasion of accessing real data. However, its poor performance also results from this limitation. By making full use of the pre-trained full-precision model $F$ to generate fake images, data synthesis has garnered more attention recently since the involvement of fake images greatly facilitates the training of quantized networks. One basic principle in data synthesis is to fit the real-data distribution, which is explored by the BNS alignment loss that aligns the batch normalization statistics (BNS) in many existing studies~\cite{zeroq,DSG,GDFQ} as:
\begin{equation}
{\cal L}_{\text{BNS}}(\tilde{\bm{I}}) = \sum_{l=1}^{L} \|\bm{\mu}'_l(\tilde{\bm{\bm{I}}})-\bm{\mu}_l^F \|^2 + \| \bm{\sigma}'_l(\tilde{\bm{\bm{I}}})-\bm{\sigma}_l^F \|_2^2,
\label{BNS loss}
\end{equation}
where $\bm{\mu}_l^F$ and $\bm{\sigma}_l^F$ are the running mean and variance stored in the $l$-th BN layer of the pre-trained full-precison network $F$, and $\bm{\mu}'_l(\tilde{\bm{\bm{I}}})$ and $\bm{\sigma}'_l(\tilde{\bm{\bm{I}}})$ denote the mean and variance of synthetic image batch $\tilde{\bm{\bm{I}}}$ in the $l$-th layer of $F$, respectively.

However, we observe that similar mean and variance do not indicate an identical data distribution. As shown in Fig.\,\ref{phenomenon:zero} and Fig.\,\ref{phenomenon:dsg}, the distributions of synthetic fake images from ZeroQ~\cite{zeroq} and DSG~\cite{DSG} differ a lot from that of the real data in Fig.\,\ref{phenomenon:real}. Particularly, a subtler class-wise distribution is overlooked since the synthesis is to fit the mean and variance of the whole dataset without any label information. The poor quality of synthetic data also results in inferior performance of 60.68\% for ZeroQ and 60.12\% for DSG on ImageNet when all layers of ResNet-18 are quantized to 4-bit as experimentally shown in Tab.\,\ref{baseline}.

Fortunately, the inception loss~\cite{Theknowledgewithin}, which first chooses an arbitrary label $y$ as a prior classification knowledge and then performs optimization to generate these label-oriented images, might be a potential method to solve this problem. It can be formulated as:
\begin{equation}
{\cal L}_{\text{IL}}(\tilde{\bm{\bm{I}}}) = ce\big(F(\tilde{\bm{\bm{I}}}), y\big),
\label{Inception loss}
\end{equation}
where $ce(\cdot, \cdot)$ represents the cross entropy and $F(\cdot)$ returns a probability distribution, \emph{i.e.}, the output of the softmax layer. Note that $F$ is fixed and the gradient will be backwarded to optimize the synthetic images $\tilde{\bm{\bm{I}}}$ for fitting the distribution of real images from class $y$. 
We can observe from Fig.\,\ref{phenomenon:zeroq+IL} and Fig.\,\ref{phenomenon:dsg+IL} that the distributions of fake data from ZeroQ and DSG become class-wise discriminative, and are more close to that of the real data after integrating the inception loss. Consequently, the performances of ZeroQ and DSG respectively increase to 63.38\% and 63.11\% in Tab.\,\ref{baseline}, well demonstrating the efficacy of incorporating prior class information in synthesizing fake images.

\begin{table}[t]
\centering
\begin{tabular}{c|c|c}
\hline
Method  & \tabincell{c}{Avg. of intra-class \\ cosine distances}      & Acc. (\%)\\ \hline \hline
full-precision & -    &  71.49 \\\hline
Real data   & 0.44      &  67.89 \\ \
ZeroQ   & -   &  60.68   \\
DSG   &  - & 60.12\\
ZeroQ+IL  & 0.17  & 63.38\\
DSG+IL & 0.19         & 63.11  \\
\textbf{IntraQ} (Ours)  &0.42  &\textbf{66.47} \\ \hline
\end{tabular}
\vspace{-0.5em}
\caption{Top-1 accuracy of 4-bit ResNet-18 fine-tuned on 5,120 fake/real images and average of intra-class cosine distances. The ``IL'' is short for inception loss.}
\vspace{-1.5em}
\label{baseline}
\end{table}

\subsection{Our Insights} 
\label{sec:Insights}

Though existing ZSQ methods benefit from the inception loss, the performance gain is limited if compared with 67.89\% of quantized ResNet-18 fine-tuned on real training data in Tab.\,\ref{baseline}. 
To dive into a deeper analysis, when looking back into Fig.\,\ref{phenomenon}, we observe that though class-wise discriminative, the synthetic data with inception loss does not well capture the intra-class heterogeneity. Images, even from the same class, often contain different contents, and thus features from the same class in Fig.\,\ref{phenomenon:real} scatter a lot. On the contrary, these in Fig.\,\ref{phenomenon:zeroq+IL} and Fig.\,\ref{phenomenon:dsg+IL} are in a dense concentration, which indicates that the synthetic images from the same class are mostly homogeneous. Quantized models fine-tuned on these homogeneous fake images fail to well generalize to the real-world test dataset featuring heterogeneity. Thus, the performance gains becomes limited.

To quantitatively measure the intra-class heterogeneity, we feed the synthetic images to the pre-trained full-precision ResNet-18 to derive their feature vectors and then calculate the cosine distances among feature data from the same class. Tab.\,\ref{baseline} displays the average cosine distance of the synthetic images. It is easy to understand that the intra-class heterogeneity can be well reflected by the cosine distance (ranging from $0$ to $2$). Note that the quantitative results for ZeroQ and DSG are not presented since the synthetic images are unlabeled when the inception loss is not applied. 
From Tab.\,\ref{baseline}, the average distance for real data is 0.44, which denotes a high degree of intra-class heterogeneity in real data. This statistical result conforms with the scattered intra-class visualization in Fig.\,\ref{phenomenon:real}. However, the average distances for ZeroQ and DSG with the inception loss are only 0.17 and 0.19, less than half of the real data. Consequently, synthetic images from the same class tend to be densely distributed in a small area as shown in Fig.\,\ref{phenomenon:zeroq+IL} and Fig.\,\ref{phenomenon:dsg+IL}. Thus, the inception loss fails to retain the intra-class heterogeneity, which however, if well addressed, might be a promise of further boosting the performance of ZSQ.

\begin{figure}[!t]
\begin{center}
\includegraphics[height=0.6\linewidth]{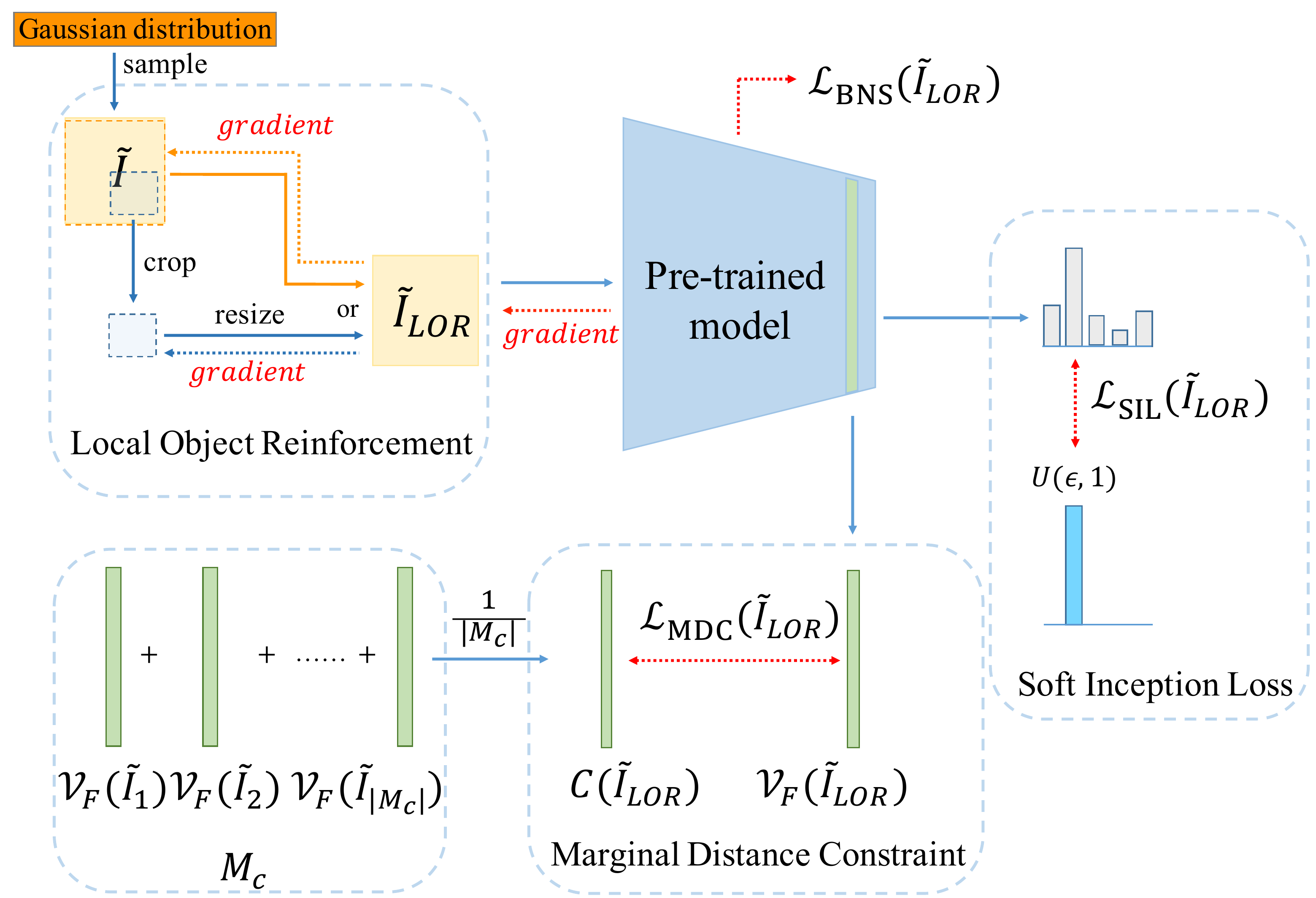}
\end{center}
\vspace{-1.0em}
\caption{The framework of our IntraQ. The local object reinforcement locates objects at different scales and positions of the synthetic image $\tilde{\bm{I}}$. The marginal distance constraint forms heterogeneous intra-class features. The soft inception injects soft label information to learn complex scenes in the synthetic images.}
\vspace{-1.5em}
\label{framework}
\end{figure}

\subsection{Our Solutions}

In what follows, we introduce our proposed IntraQ to learn synthetic images with intra-class heterogeneity. As shown in Fig.\,\ref{framework}, the core contributions of our IntraQ are three folds: a local object reinforcement, a marginal distance constraint, and a soft inception loss. We elaborate on the details in the following subsections.

\subsubsection{Local Object Reinforcement}
\label{sec:Local Object Reinforcement}

Our first step to retain the intra-class heterogeneity resorts to enhancing the synthetic images before feeding them to the pre-trained full-precision model $F$.
Our motives lie in the fact that the objects of interest the model is expected to learn are not always at the same scale or position in the images. 
%
%
Thus, it is natural to expect the synthetic images to contain these informative contents at different scales or positions.
However, earlier methods only focus on optimizing a complete image during the whole process of data generation given prior labels from the inception loss. Thus, the synthetic images tend to have the target objects all at the scale of covering up the whole images, and these synthetic images with the same prior label become very similar, which thus fails to retain the intra-class heterogeneity.

Motivated by the above analysis, we propose to locate the target objects at different scales and positions of the synthetic images. Specifically, for each synthetic image, instead of directly feeding the whole image to the pre-trained full-precision model $F$ for optimization, we choose to randomly crop a patch of the image with a probability of $p$. For each cropping, its scaling rate is sampled from a uniform distribution $U(\eta, 1)$ where $\eta$ is a hyper-parameter controlling the minimum scaling rate of the cropped patch. Thus, the input of the pre-trained full-precision $F$ after our local object reinforcement becomes:
\begin{equation}
    \tilde{\bm{I}}_{LOR} = 
    \begin{cases}
    resize\big(crop_{\eta}(\tilde{\bm{I}})\big) \;\; {\text{with probability of $p$}}\\
    \tilde{\bm{I}} \;\; {\text{with probability of $1-p$}},
    \end{cases}
\label{CR}
\end{equation}
where $crop_{\eta}(\cdot)$ randomly crops a patch from its input with a scaling rate sampled from a uniform distribution of $U(\eta, 1)$ where $\eta$ is a pre-defined parameter, and $resize(\cdot)$ resizes its input to the size of the original synthetic image $\tilde{\bm{I}}$. The $p = 50\%$ is observed to perform best.

To stress, traditional cropping in data augmentation is to discard irrelevant contents and remain the pleasing portions of the image to enhance the overall composition. In contrast, our image cropping is to synthesize fake images with target objects at different scales and positions. As shown in Fig.\,\ref{framework}, if cropped, the gradient is only backward to update the local cropped patch. Given a prior label, the cropping is applied at different positions and scales, and thus the synthesized images are no longer similar, which further retains the intra-class heterogeneity.


\subsubsection{Marginal Distance Constraint}
\label{sec:Marginal Distance Constraint}

With the enhanced synthetic images as inputs, we can extract their feature vectors from the pre-trained full-precision network $F$. To correctly classify all the classes simultaneously, we expect $F$ to form class-related features with large intra-class discrimination. To well generalize the quantized model to the real-world test dataset, we also expect $F$ to form heterogeneous intra-class features. To achieve this, we further devise the following marginal distance constraint as a supervisory signal to guide the feature learning:
%
%
\begin{equation}
\begin{split}
{\cal L}_{\text{MDC}}({\tilde{\bm{I}}}_{LOR}) &=  max\Big(\lambda_l - cos\big(\mathcal{V}_F(\tilde{\bm{I}}_{LOR}), \mathcal{C}(\tilde{\bm{I}}_{LOR})\big), 0\Big) \\ & + max\Big(cos\big(\mathcal{V}_F(\tilde{\bm{I}}_{LOR}), \mathcal{C}(\tilde{\bm{I}}_{LOR})\big) - \lambda_u, 0\Big),
\end{split}
\label{MDC loss}
\end{equation}
where $cos(\cdot, \cdot)$ returns the cosine distance of its two inputs in comparison and  $\mathcal{V}_F(\cdot)$ returns the feature vector extracted by the pre-trained full-precision $F$, and $\mathcal{C}(\tilde{\bm{I}}_{LOR})$ returns the class center of $\tilde{\bm{I}}_{LOR}$. Assuming the label of $\tilde{\bm{I}}_{LOR}$ is $c$ and $M_c$ is a collection of the previously generated synthetic images belonging to class $c$, we define the class center as the mean feature vector of all synthetic images in $M_c$:
\begin{equation}
C(\tilde{\bm{I}}_{LOR}) = \frac{1}{|M_c|} \sum_{i=1}^{|M_c|} \mathcal{V}_F(\tilde{\bm{I}}_i), \:\:\tilde{\bm{I}}_i \in M_c.
\label{class center} 
\end{equation}

The $\lambda_l$ and $\lambda_u$ in Eq.\,(\ref{MDC loss}) are two hyper-parameters to control the lower and upper bounds of the cosine distance between $\tilde{\bm{I}}_{LOR}$ and its class center. Detailedly, Eq.\,(\ref{MDC loss}) requires the distance larger than a margin $\lambda_l$, but smaller than a margin $\lambda_u$. The upper bound $\lambda_u$ encourages features extracted from fake images of the same class to be similar, which brings about correct classification. The low bound $\lambda_l$ avoids learning features concentrated on a dense area and thus can effectively preserve the intra-class heterogeneity, which ensures the generalization ability when the quantized model is employed on real-world test data.

\subsubsection{Soft Inception Loss}
\label{sec:Soft Inception loss}

The inception loss of Eq.\,(\ref{Inception loss}) is to inject prior label knowledge into the synthetic images. To fulfill this goal, the loss essentially drives the gradients to optimize the synthetic images until the output of the pre-trained network $F$ exactly matches the one-hot label. However, image contents are often overlapped even though they are grouped into different classes. One-hot labels do not represent soft decision boundaries among different objects, and hence the synthetic images trained on them are prone to overfitting to a fixed object. These images tend to be ``easy'' and do not well acquire the complex scenes within the contents. Consequently, existing methods embedded with the inception loss fail to retain the intra-class heterogeneity as shown in Fig.\,\ref{phenomenon}.

Reflecting on this, we consider soft labels as a regularization that has the potential to tell a model more about the meaning of each synthetic image. To be specific, given the enhanced synthetic image $\tilde{\bm{I}}_{LOR}$ with its prior label $y = c$, we devise the following soft inception loss:
\begin{equation}
{\cal L}_{\text{SIL}}({\tilde{\bm{I}}}_{LOR}) = mse\big(F(\tilde{\bm{\bm{I}}}_{LOR})_c, U(\epsilon, 1) \big),
\label{Soft Inception loss}
\end{equation}
where $\epsilon$ is a pre-defined parameter to control the softness of the label vector. Recall that $F(\cdot)$ returns the output of the softmax layer as stated in Sec.\,\ref{data synthesis}. Herein, $F(\cdot)_c$ indicates the $c$-th element of $F(\cdot)$. The $mse(\cdot, \cdot)$ computes the mean squared error between its two inputs.

Our soft inception loss requires the prediction probability of each synthetic image to match a soft label randomly sampled from a uniform distribution of $U(\epsilon, 1)$ instead of the hard one-hot form. Consequently, the synthetic images no longer overfit to a fixed object labeled with $y = c$, and more complex scenes are excavated, which further benefits the desired property of intra-class heterogeneity.

\subsection{Training Process}
\label{sec:training process}

Our learning of a quantized network consists of two parts including a data generation for fake images and fine-tuning of the quantized network upon the fake images.

\subsubsection{Data Generation}

We start with random input data $\tilde{\bm{I}}$ drawn from a standard Gaussian distribution. Our data generation aims to optimize $\tilde{\bm{I}}$, such that the distribution of the fake data can match that of real data in particular with the intra-class heterogeneity.
To this end, as shown in Fig.\,\ref{framework}, we first apply our local object reinforcement detailed in Sec.\,\ref{sec:Local Object Reinforcement} to derive $\tilde{\bm{I}}_{LOR}$. Then, we feed $\tilde{\bm{I}}_{LOR}$ to the pre-trained full-precision network $F$ to compute the BNS alignment loss of Eq.\,(\ref{BNS loss}) and our proposed marginal distance constraint of Eq.\,(\ref{MDC loss}). Also, we replace the traditional inception of Eq.\,(\ref{Inception loss}) with our proposed soft inception loss of Eq.\,(\ref{Soft Inception loss}). As such, our final loss for data generation can be obtained as:
\begin{equation}
{\cal L}(\tilde{\bm{I}}_{LOR}) = {\cal L}_{\text{BNS}}({\tilde{\bm{I}}_{LOR}}) + {\cal L}_{\text{MDC}}({\tilde{\bm{I}}_{LOR}}) + {\cal L}_{\text{SIL}}({\tilde{\bm{I}}_{LOR}}).
\label{data loss}
\end{equation}

\subsubsection{Network Fine-Tuning}
With our synthetic fake images $\tilde{\bm{I}}$, we apply them to fine-tune the quantized network $Q$ with the cross-entropy loss:
\begin{equation}
{\cal L}^Q_{\text{CE}} = ce\big(Q(\tilde{\bm{I}}), y\big).
\label{ce_q}
\end{equation}
Following~\cite{GDFQ}, we also transfer the output of $F$ to $Q$ as:
\begin{equation}
{\cal L}_{\text{KD}}^Q = kl\big(Q(\tilde{\bm{I}}), F(\tilde{\bm{I}})\big),
\label{kd_q}
\end{equation}
where $kl(\cdot, \cdot)$ computes the Kullback-Leibler distance between its two inputs. Thus, the overall loss for fine-tuning the quantized network $Q$ can be summarized as:
\begin{equation}
{\cal L}^Q = {\cal L}_{\text{CE}}^Q+\alpha \cdot {\cal L}_{\text{KD}}^Q,
\label{loss_q}
\end{equation}
where $\alpha$ balances the importance of ${\cal L}_{\text{CE}}^Q$ and ${\cal L}_{\text{KD}}^Q$.

In Tab.\,\ref{baseline}, our IntraQ results in an average intra-class cosine distance of 0.42, very close to 0.44 of the real data. Besides, the visualization results show that the distribution of our synthetic images (Fig.\,\ref{phenomenon:ours}) also approximates that of real data. Moreover, IntraQ obtains 66.47\% in the top-1 accuracy, over 3.0\% increases compared with ZeroQ and DSG integrated with the inception loss, as illustrated in Tab.\,\ref{baseline}.

\begin{table}[!t]
\centering
\begin{subtable}{\linewidth}\centering
\begin{tabular}{c|c|c|c}
\hline 
Bit-width & Method  &Generator   & Acc. (\%)\\ \hline \hline
& full-precision      &  - & 94.03  \\ \hline
& Real data &  -  & 91.52 \\
& GDFQ   & \Checkmark &  90.25 \\
& ZeroQ   & \XSolidBrush  &   84.68  \\
& DSG   & \XSolidBrush &  88.74 \\
& ZeroQ+IL & \XSolidBrush  &  89.66\\
& DSG+IL   & \XSolidBrush & 88.93 \\
& GZNQ & \XSolidBrush  &  91.30 \\
\multirow{-8}{*}{W4A4} &  \textbf{IntraQ} (Ours)  &  \XSolidBrush   & \textbf{91.49} \\ \hline
& Real data      & - &87.94 \\
& GDFQ   & \Checkmark &   71.10 \\
& ZeroQ   & \XSolidBrush  &   29.32  \\
& DSG   & \XSolidBrush & 32.90 \\
& ZeroQ+IL & \XSolidBrush  & 69.53 \\
& DSG+IL   & \XSolidBrush &  48.99 \\
\multirow{-7}{*}{W3A3} &  \textbf{IntraQ} (Ours)   &  \XSolidBrush   & \textbf{77.07} \\ \hline
\end{tabular}
\captionsetup{font={normalsize}}
\subcaption{CIFAR-10}\label{comparsion:cifar10}
\end{subtable}

\begin{subtable}{\linewidth}\centering
\begin{tabular}{c|c|c|c}
\hline 
Bit-width & Method   & Generator   & Acc. (\%)\\ \hline \hline
& full-precision      &  - &  70.33 \\ \hline
& Real data &  -  &66.80 \\
& GDFQ   &\Checkmark & 63.58 \\
& DSG   & \XSolidBrush &  62.36 \\
& ZeroQ   &    \XSolidBrush  &  58.42   \\
& DSG+IL   & \XSolidBrush &  62.62 \\
& ZeroQ+IL &  \XSolidBrush  & 63.97 \\
& GZNQ &  \XSolidBrush  & 64.37 \\
\multirow{-8}{*}{W4A4} & \textbf{IntraQ} (Ours)    &  \XSolidBrush   & \textbf{64.98} \\ \hline
& Real data &  -  &56.26 \\
& GDFQ   & \Checkmark & 43.87 \\
& DSG   & \XSolidBrush & 25.48 \\
& ZeroQ   &    \XSolidBrush  & 15.38  \\
& DSG+IL   & \XSolidBrush & 43.42  \\
& ZeroQ+IL &  \XSolidBrush  & 26.35 \\
\multirow{-7}{*}{W3A3} & \textbf{IntraQ} (Ours)  &  \XSolidBrush   & \textbf{48.25} \\ \hline
\end{tabular}
\captionsetup{font={normalsize}}
\subcaption{CIFAR-100}\label{comparsion:cifar100}
\end{subtable}
\vspace{-0.5em}
\caption{Results of ResNet-20 on CIFAR-10/100. WBAB indicates the weights and activations are quantized to B-bit.}
\vspace{-1.0em}
\label{comparsion:cifar10/100}
\end{table}

\begin{table}[!t]
\centering
\begin{subtable}{\linewidth}\centering
\begin{tabular}{c|c|c|c}
\hline 
Bit-width  & Method  & Generator   & Acc. (\%)\\ \hline \hline
& full-precision      &  - &  71.47 \\ \hline
& Real data &  -  & 70.31\\
& GDFQ   & \Checkmark & 66.82 \\
& DSG   & \XSolidBrush &  69.53 \\
& ZeroQ   &    \XSolidBrush  &  69.65   \\
& DSG+IL   & \XSolidBrush & 69.53  \\
& ZeroQ+IL &  \XSolidBrush  & 69.72\\
\multirow{-7}{*}{W5A5} &   \textbf{IntraQ} (Ours)     &  \XSolidBrush   & \textbf{69.94} \\ \hline
& Real data &  -  & 67.89\\
& GDFQ   & \Checkmark & 60.60 \\
& DSG+G   & \Checkmark & 61.58 \\
& ZeroQ   &    \XSolidBrush  &  60.68   \\
& DSG   & \XSolidBrush & 60.12 \\
& ZeroQ+IL &  \XSolidBrush  & 63.38\\
& DSG+IL   & \XSolidBrush & 63.11  \\
& GZNQ &  \XSolidBrush  & 64.50\\
\multirow{-8}{*}{W4A4} & \textbf{IntraQ} (Ours)            &  \XSolidBrush   & \textbf{66.47} \\ \hline
\end{tabular}
\captionsetup{font={normalsize}}
\end{subtable}
\vspace{-0.5em}
\caption{Results of ResNet-18 on ImageNet. WBAB indicates the weights and activations are quantized to B-bit.}
\vspace{-1.0em}
\label{comparsion:resnet}
\end{table}

\section{Experiments}
\subsection{Implementation Details}
\label{sec:Implementation}

We report top-1 accuracy on the validation set of CIFAR-10/100~\cite{cifar} and ImageNet~\cite{russakovsky2015imagenet}.
The quantized networks include ResNet-20~\cite{he2016deep} for CIFAR-10/100, ResNet-18~\cite{he2016deep}, MobileNetV1~\cite{howard2017mobilenets} and MobileNetV2~\cite{sandler2018mobilenetv2} for ImageNet. All experiments are implemented with Pytorch~\cite{paszke2019pytorch}.

For our data generation, the Adam~\cite{kingma2014adam} is adopted with a momentum of 0.9 and an initial learning rate of 0.5. We update the synthetic images for 1,000 iterations and decay the learning rate by a factor of 0.1 each time the data generation loss of Eq.\,(\ref{data loss}) stops decreasing for 50 iterations. The batch size of synthetic images is set as 256.
There are four hyper-parameters in our data generation, including $\eta$ in Eq.\,(\ref{CR}), $\lambda_l$ and $\lambda_u$ in Eq.\,(\ref{MDC loss}), and $\epsilon$ in Eq.\,(\ref{Soft Inception loss}). They are respectively set to 0.5, 0.05, 0.8, and 0.9 on CIFAR-10; 0.5, 0.02, 1.0, and 0.6 on CIFAR-100; 0.5, 0.3 0.8, and 0.9 on ImageNet. 
As for ZeroQ+IL and DSG+IL, we implement the experiments based on their open-source code and use the same training configurations as ours.

For all datasets, we generate 5,120 synthetic images to fine-tune the quantized model using SGD with Nesterov~\cite{nesterov1983method}. We set the weight decay as 10$^{-4}$ and a total of 150 fine-tuning epochs are given. The batch size for fine-tuning is 256 for CIFAR-10/100 and 16 for ImageNet. Besides, CIFAR-10/100 is in configuration with an initial learning rate of 10$^{-4}$ while it is 10$^{-6}$ for ImageNet. Both learning rates are decayed by 0.1 every 100 fine-tuning epochs. The hyper-parameter in our network fine-tuning is $\alpha$ in Eq.\,(\ref{loss_q}) which is always set to 20.


\begin{table}[!t]
\centering
\begin{subtable}{\linewidth}\centering
\begin{tabular}{c|c|c|c}
\hline 
Bit-width & Method   &Generator   & Acc. (\%)\\ \hline \hline
& full-precision      &  - & 73.39  \\ \hline
& Real data   & - & 69.87 \\
& GDFQ   & \Checkmark &   59.76 \\
& ZeroQ   &    \XSolidBrush  &   61.95  \\
& DSG   & \XSolidBrush & 64.18 \\
& ZeroQ+IL &  \XSolidBrush  & 67.11 \\
& DSG+IL   & \XSolidBrush &  66.61 \\
\multirow{-6}{*}{W5A5} & \textbf{IntraQ} (Ours)     &  \XSolidBrush   &  \textbf{68.17}\\ \hline
& Real data  & - &  59.66 \\
& GDFQ   & \Checkmark &  28.64 \\
& ZeroQ   &    \XSolidBrush  &    20.96 \\
& DSG   & \XSolidBrush & 21.14 \\
& ZeroQ+IL &  \XSolidBrush  & 25.43 \\
& DSG+IL   & \XSolidBrush &  42.19 \\
\multirow{-7}{*}{W4A4} &   \textbf{IntraQ} (Ours)        &  \XSolidBrush   &  \textbf{51.36} \\ \hline
\end{tabular}
\captionsetup{font={normalsize}}
\subcaption{MobileNetV1}\label{comparsion:mv1}
\end{subtable}

\begin{subtable}{\linewidth}\centering
\begin{tabular}{c|c|c|c}
\hline 
Bit-width & Method    &Generator   & Acc. (\%)\\ \hline \hline
& full-precision      &  - &  73.03 \\ \hline
& Real data   & - & 72.01 \\
& GDFQ   & \Checkmark & 68.14 \\
& ZeroQ   &    \XSolidBrush  &   70.88  \\
& DSG   & \XSolidBrush & 70.85\\
& ZeroQ+IL &  \XSolidBrush  & 70.95 \\
& DSG+IL   & \XSolidBrush & 70.87  \\
\multirow{-7}{*}{W5A5} & \textbf{IntraQ} (Ours)    &  \XSolidBrush   & \textbf{71.28} \\ \hline
& Real data      & - & 67.90  \\
& GDFQ   & \Checkmark & 51.30 \\
& DSG+G   & \Checkmark & 54.66 \\
& GZNQ & \XSolidBrush & 53.53 \\
& ZeroQ   &    \XSolidBrush  &    59.39  \\
& DSG   & \XSolidBrush &  59.04 \\
& ZeroQ+IL &  \XSolidBrush  & 60.15\\
& DSG+IL   & \XSolidBrush & 60.45  \\
\multirow{-8}{*}{W4A4} &      \textbf{IntraQ} (Ours)    &  \XSolidBrush   & \textbf{65.10} \\ \hline
\end{tabular}
\captionsetup{font={normalsize}}
\subcaption{MobileNetV2}\label{comparsion:mv2}
\end{subtable}
\vspace{-0.5em}
\caption{Results of MobileNetV1/V2 on ImageNet. WBAB indicates the weights and activations are quantized to B-bit.}
\vspace{-1.0em}
\label{comparsion:mobilenet}
\end{table}

\begin{figure*}[!t]
  \centering
  \begin{subfigure}{0.18\linewidth}
    \includegraphics[width=\linewidth]{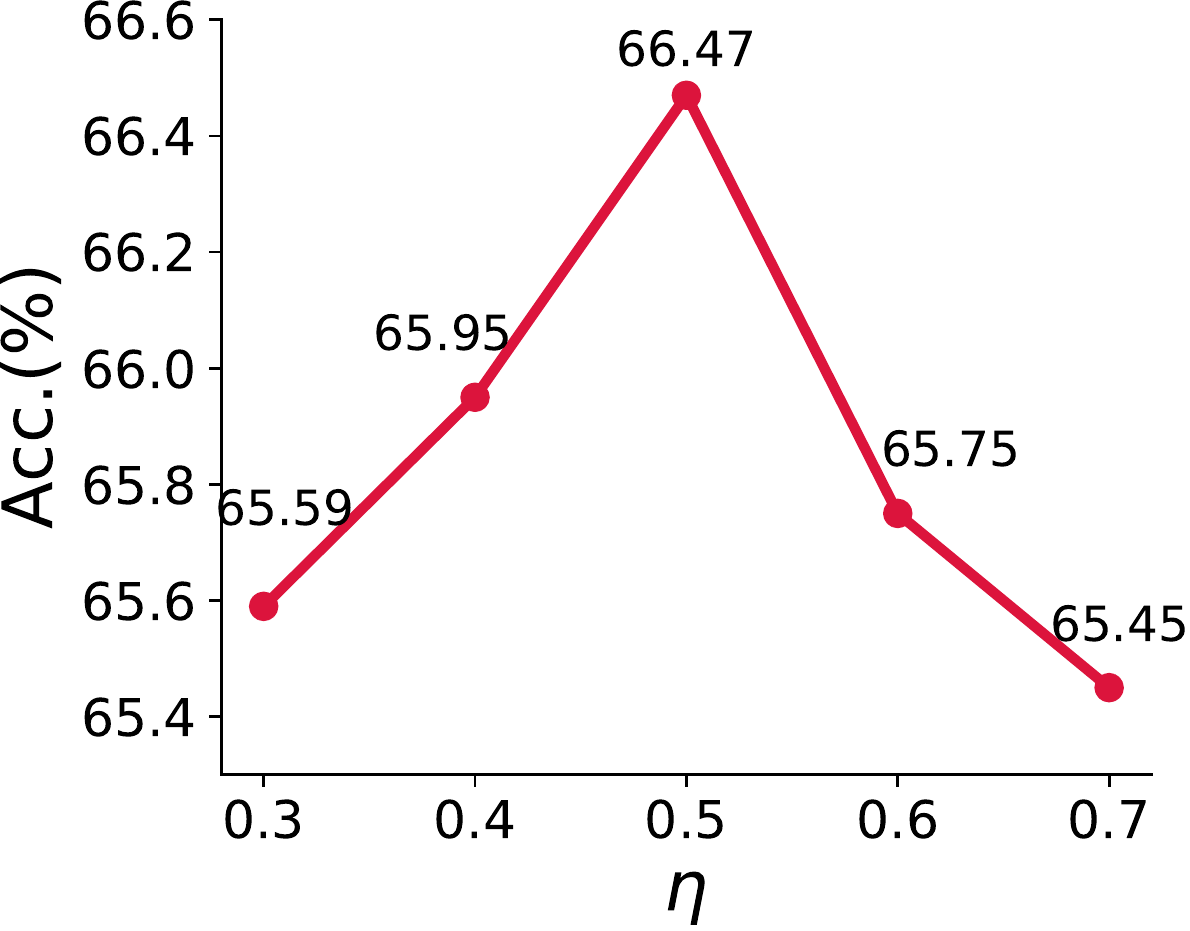}
    \caption{}
    \label{ablation:eta}
  \end{subfigure}
  \hspace{0.13em}
  \begin{subfigure}{0.18\linewidth}
    \includegraphics[width=\linewidth]{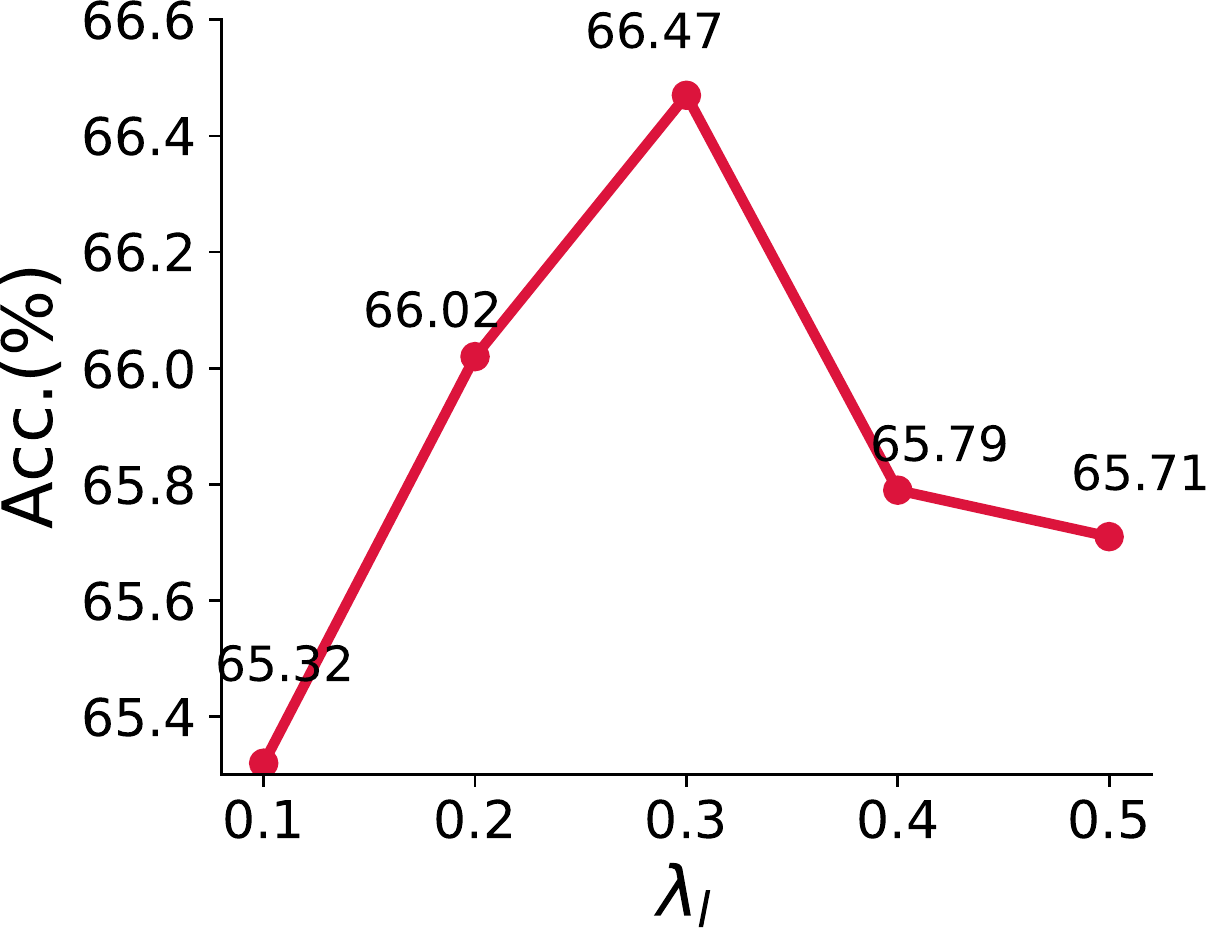}
    \caption{}
    \label{ablation:lambda_l}
  \end{subfigure}
  \hspace{0.13em}
  \begin{subfigure}{0.18\linewidth}
    \includegraphics[width=\linewidth]{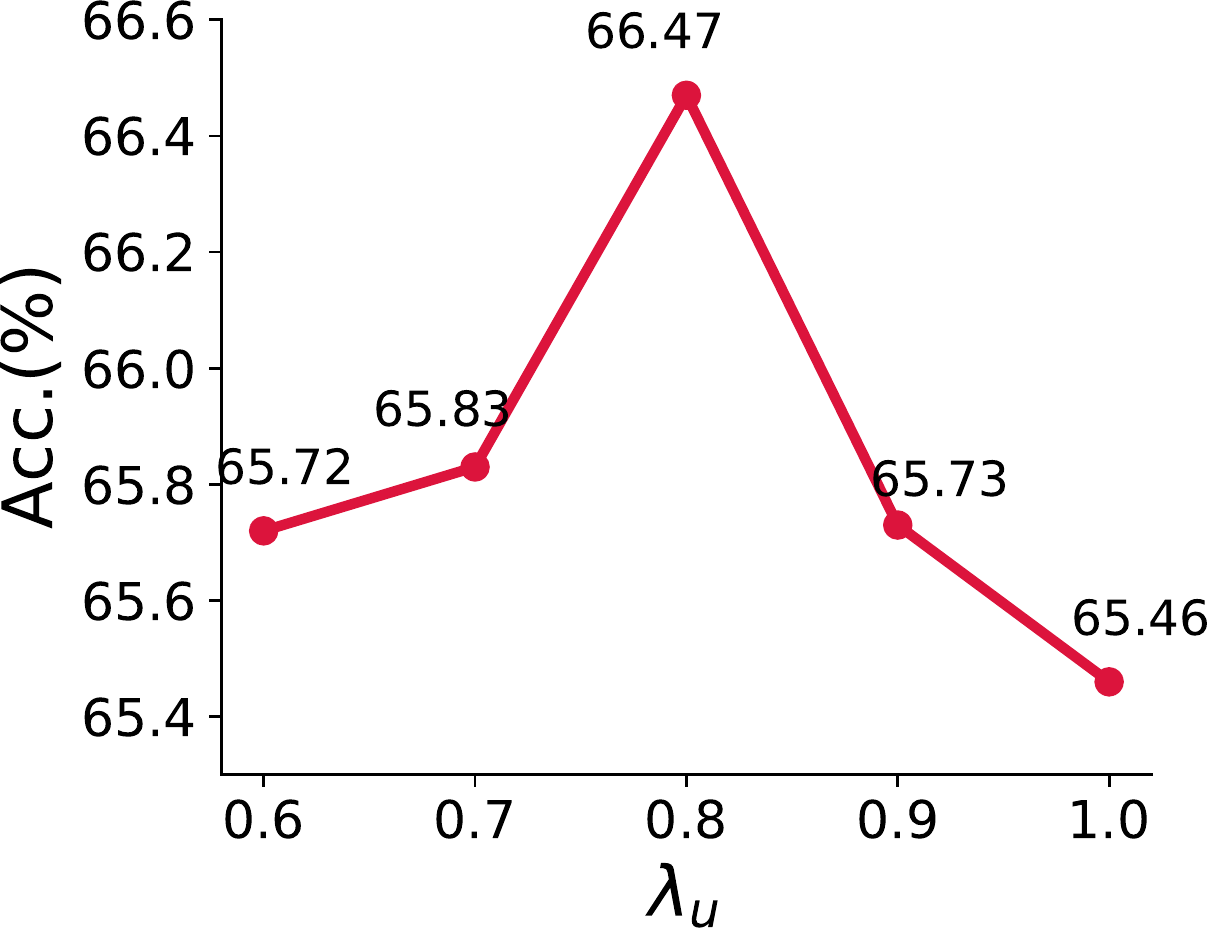}
    \caption{}
    \label{ablation:lambda_u}
  \end{subfigure}
  \hspace{0.13em}
  \begin{subfigure}{0.18\linewidth}
    \includegraphics[width=\linewidth]{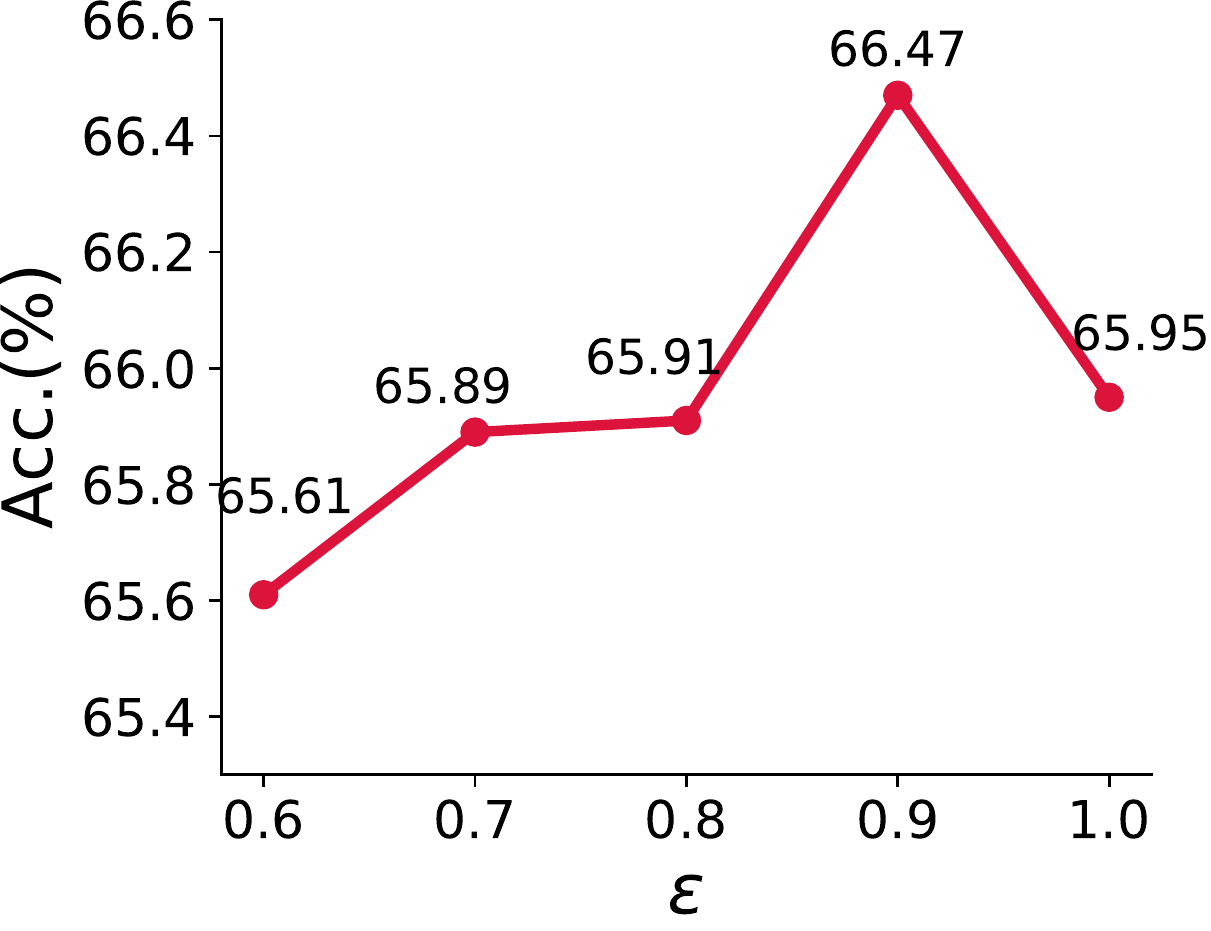}
    \caption{}
    \label{ablation:epsilon}
  \end{subfigure}
  \hspace{0.13em}
  \begin{subfigure}{0.18\linewidth}
    \includegraphics[width=\linewidth]{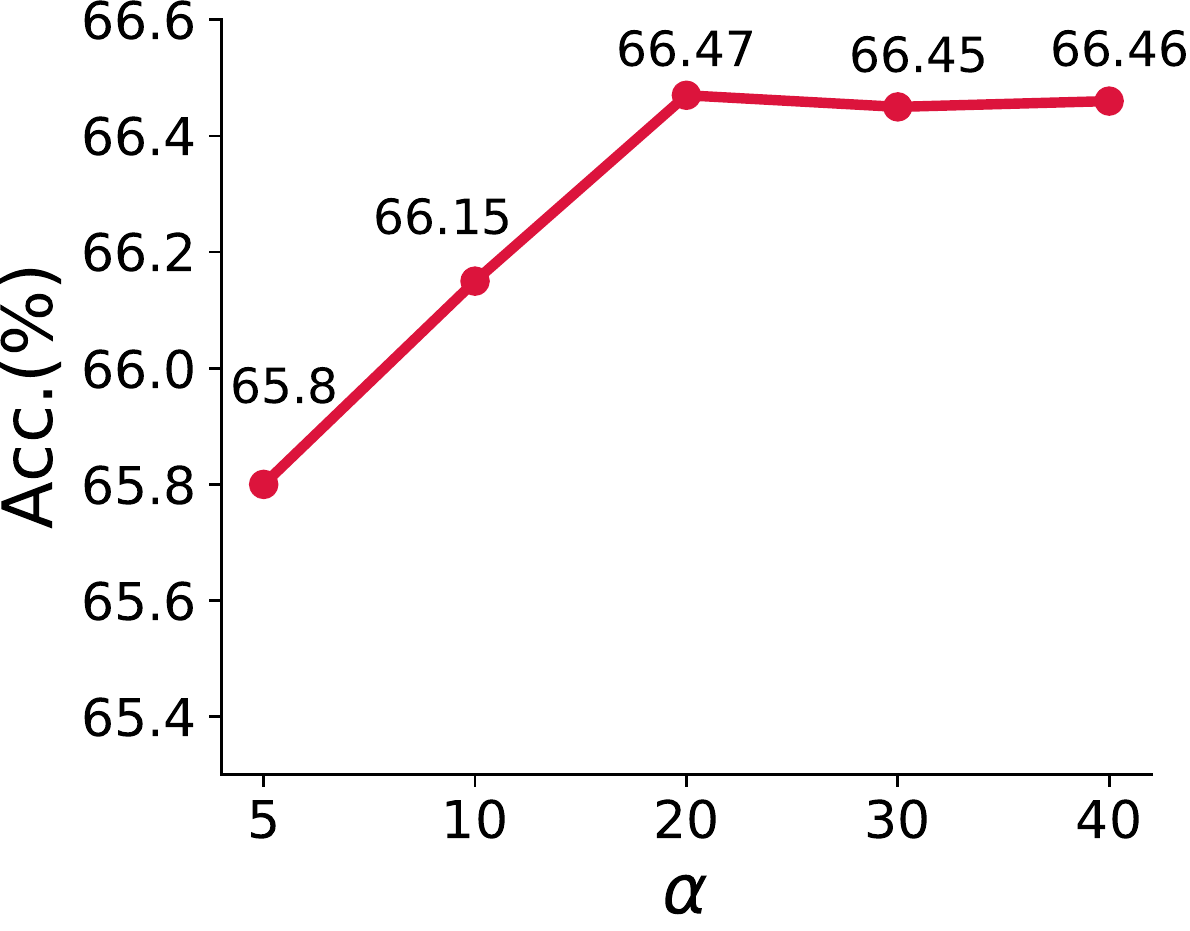}
    \caption{}
    \label{ablation:full-sized image}
  \end{subfigure}
  \vspace{-0.5em}
  \caption{Influence of the hyper-parameters to the top-1 accuracy of 4-bit ResNet-18 on ImageNet.}
  \label{ablation:hyperparameters}
  \vspace{-1.0em}
\end{figure*}
\subsection{Performance Comparison}

\subsubsection{CIFAR-10/100}

We first analyze the performance on CIFAR-10/100, comparing it against the popular ZSQ methods including GDFQ~\cite{GDFQ}, ZeroQ~\cite{zeroq}, DSG~\cite{DSG} and GZNQ~\cite{VAEGenerate}. To demonstrate the efficacy of our IntraQ, we quantize all layers of ResNet-20 including the first and last layers to the ultra-low precisions of 4-bit and 3-bit since CIFAR-10/100 are relative simple datasets and high performance can be easily reached if larger quantization bits are given.

From Tab.\,\ref{comparsion:cifar10/100}, we can see that our IntraQ consistently outperforms the compared methods on both CIFAR-10 and CIFAR-100. Specifically, compared to the advanced generator-based GDFQ, our IntraQ increases the top-1 accuracy of 3-bit quantized models by 5.97\% on CIFAR-10 and 4.38\% on CIFAR-100. Similar results can be observed in 4-bit quantization as well. In particular, compared with GZNQ~\cite{VAEGenerate} that requires 50,000 synthetic images to obtain accuracies of 91.30\% and 64.37\% on CIFAR-10 and CIFAR-100, the proposed IntraQ reaches the higher performance of 91.49\% and 64.98\% using only 5,120 synthetic images, well demonstrating the superiority of exploiting the intra-class heterogeneity in the synthetic fake images.

\subsubsection{ImageNet}

We further compare with the competitors on the large-scale ImageNet. The quantized networks include ResNet-18 and MobileNetV1/V2. Similar to CIFAR-10/100, we quantize all layers of the networks including the first and last layers. Differently, we display the results of 5-bit and 4-bit due to the large scale of ImageNet.

\textbf{ResNet-18}. 
Tab.\,\ref{comparsion:resnet} shows the experimental results of ResNet-18. In the case of 5-bit, our IntraQ slightly outperforms the existing method ZeroQ with inception loss (69.94\% \emph{vs}. 69.72\%). When it comes to 4-bit, a noticeable increase is observed from the proposed method. Detailedly, GZNQ obtains a limited accuracy of 64.50\% using a total of 100,000 synthetic images. On the contrary, our IntraQ retains a high performance of 66.47\% using only 5,120 synthetic images for fine-tuning the quantized ResNet-18, leading to 1.97\% accuracy increases.


\begin{table}[!t]
\centering
\begin{tabular}{ccc|c}
\hline
 LOR    & MDC &  SIL &   Acc. (\%)\\ \hline \hline
 &  ZeroQ+IL &  &  63.38 \\\hline
\Checkmark &  & & 66.14 \\
 & \Checkmark & & 63.77 \\
 &  & \Checkmark &   63.60 \\
      & \Checkmark & \Checkmark & 64.05 \\
      \Checkmark     & \Checkmark & & 66.32  \\
 \Checkmark     & & \Checkmark & 66.30 \\
 \Checkmark  & \Checkmark & \Checkmark  &  \textbf{66.47}   \\\hline
\end{tabular}
\vspace{-0.5em}
\caption{Ablations on different components of our IntraQ. ``LOR'' represents the local object reinforcement, ``MDC'' represents the marginal distance constraint, and ``SIL'' represents the soft inception loss. We report the top-1 accuracy of 4-bit ResNet on ImageNet.}
\vspace{-1.0em}
\label{ablation:components}
\end{table}

\textbf{MobileNetV1/V2}.
In Tab.\,\ref{comparsion:mobilenet}, compared with the state-of-the-art ZeroQ+IL in 5-bit and DSG+IL in 4-bit, the proposed IntraQ still maintains the best performance in quantizing the light-weight MobileNetV1/V2. The supreme performance is in particular obvious in the case of lower 4-bit. For instance, our IntraQ obtains 9.17\% accuracy improvements compared with the advanced DSG+IL when all layers of MobileNetV1 are represented in a 4-bit form. These results again demonstrate the effectiveness of our synthetic images for ZSQ and also verify the correctness of our motive to excavate the intra-class heterogeneity.


\subsection{Ablation Study}

In this section, we conduct ablation studies of the hyper-parameters and different components of our IntraQ. All experiments are conducted by quantizing all layers of ResNet-18 to 4-bit on ImageNet. The top-1 accuracy is reported.

\textbf{Hyper-parameters}. 
We first display the influence of different hyper-parameters including $\eta$ in Eq.\,(\ref{CR}), $\lambda_l$ and $\lambda_u$ in Eq.\,(\ref{MDC loss}), $\epsilon$ in Eq.\,(\ref{Soft Inception loss}), and $\alpha$ in Eq.\,(\ref{loss_q}). From Fig.\,\ref{ablation:hyperparameters}, we can see that the optimal results of these parameters are $\eta$ = 0.5, $\lambda_l$ = 0.3, $\lambda_u$ = 0.8, $\epsilon$ = 0.9, and $\eta$ = 20. To avoid a cumbersome search, we use these configurations for all experiments on ImageNet. Though these results might not be optimal for all networks, we find they already bring the best performance in comparison with existing methods.
Similar experiments can be conducted to find out the optimal values of these parameters on other datasets, which have been listed in Sec.\,\ref{sec:Implementation}.

\textbf{Components}.
We further study the effectiveness of our proposed local object reinforcement in Sec.\,\ref{sec:Local Object Reinforcement}, marginal distance constraint in Sec.\,\ref{sec:Marginal Distance Constraint}, and soft inception loss in Sec.\,\ref{sec:Soft Inception loss}. Tab.\,\ref{ablation:components} shows the experimental results. Note that ZeroQ+IL can serve as a baseline since it uses BNS alignment loss in Eq.\,(\ref{BNS loss}) and inception loss in Eq.\,(\ref{Inception loss}). As can be seen, when the three strategies are individually added to synthesize fake images, the accuracy increases compared with the baseline of ZeroQ+IL. Among them, the local object reinforcement significantly boosts the baseline from 63.38\% to 66.14\%. This inspires us of the importance of synthesizing images with objects in different scales and positions in order to retain the intra-class heterogeneity. Furthermore, the performance continues to increase if two of them are used together. When all of the three strategies are applied, the best performance of 66.47\% can be obtained.

\section{Limitations}


Though the proposed IntraQ improves the accuracy of existing ZSQ methods by a large margin, its performance still degrades a lot if compared with the results of real data. Thus, how to further improve the quality of fake data remains to be investigated in our future work.
Due to our limited hardware resources, we are unable to perform our IntraQ on other computer vision tasks (\emph{e.g.}, detection). It is unclear whether the intra-class heterogeneity can still be observed, thus the applicability of our InterQ on other tasks remains an open issue. More efforts are required to address this problem in our near future.

\section{Conclusion}
In this paper, we investigate optimizing synthetic images for zero-shot quantization (ZSQ). We discover a non-ignorable phenomenon of intra-class heterogeneity in real data. To retain this property in the synthetic images for better performance, we propose a novel ZSQ method, called IntraQ. To that effect, our innovations are three folds including a local object reinforcement, a marginal distance constraint, and a soft inception loss. The local object reinforcement locates the target objects at different scales and positions of the synthetic images to avoid producing similar images. The marginal distance constraint is applied to prevent image features from being concentrated together. The soft inception loss considers a soft label as prior knowledge to excavate more complex scenes within the synthetic images. With our innovations, the synthetic images are demonstrated to be heterogeneous within each class and the quantized models fine-tuned on these images are experimentally shown to be superior in performance.

\textbf{Acknowledgement}. This work is supported by the National Science Fund for Distinguished Young Scholars (No.62025603), the National Natural Science Foundation of China (No.U1705262, No.62072386, No.62072387, No.62072389, No.62002305, No.61772443, No.61802324 and No.61702136), Guangdong Basic and Applied Basic Research Foundation (No.2019B1515120049) and the Fundamental Research Funds for the central universities (No.20720200077, No.20720200090 and No.20720200091).


{\small
\bibliographystyle{ieee_fullname}
\bibliography{egbib}
}

\clearpage

\section*{Appendix \label{appendix}}

\section{MDC \textit{vs}. ADI}

\begin{figure}[!ht]
\centering
\begin{subfigure}{0.45\linewidth}
    \includegraphics[width=\linewidth]{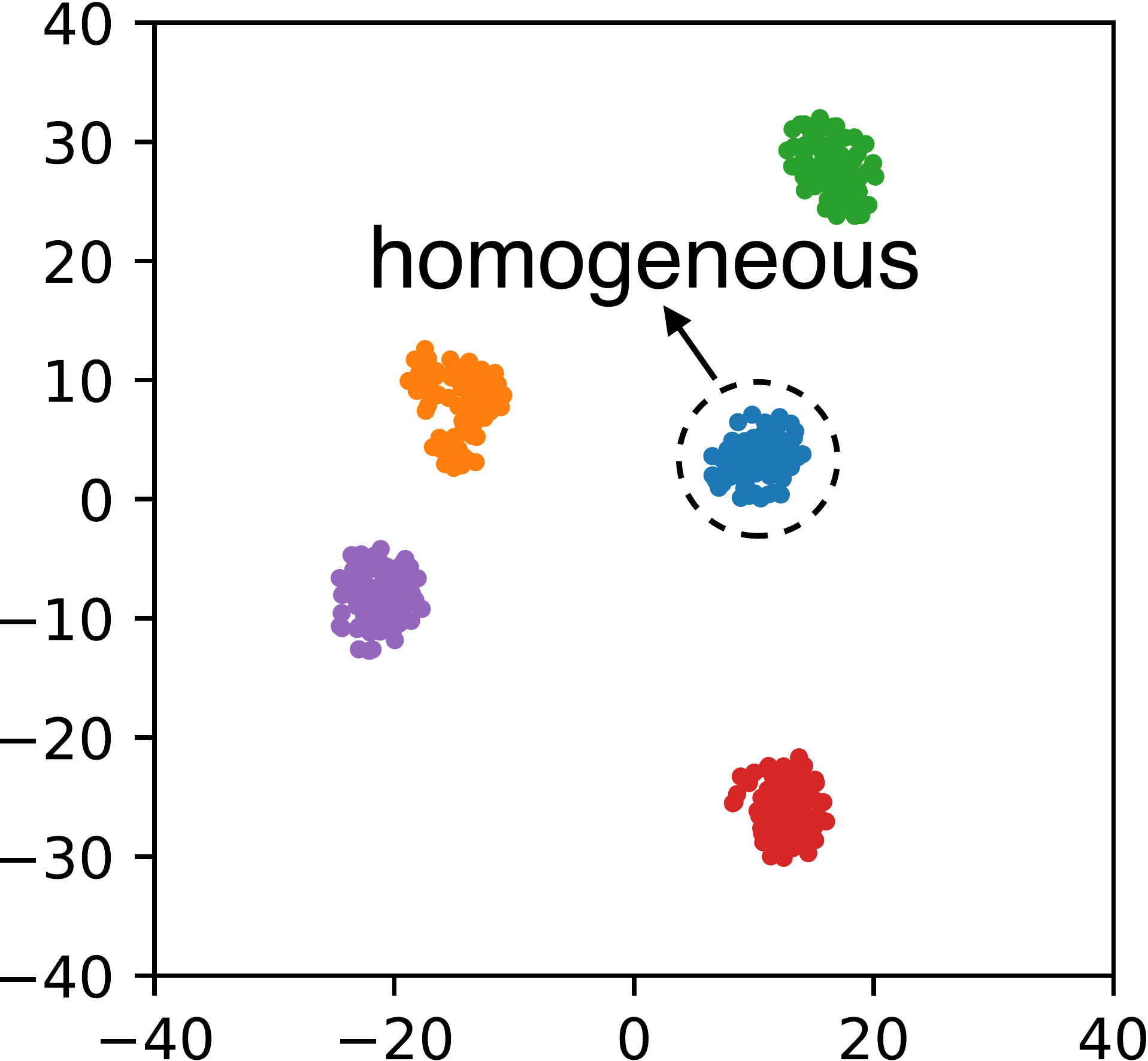}
    \caption{ADI}
    \label{visualization:adi}
\end{subfigure}
\hspace{3mm}
\begin{subfigure}{0.45\linewidth}
    \includegraphics[width=\linewidth]{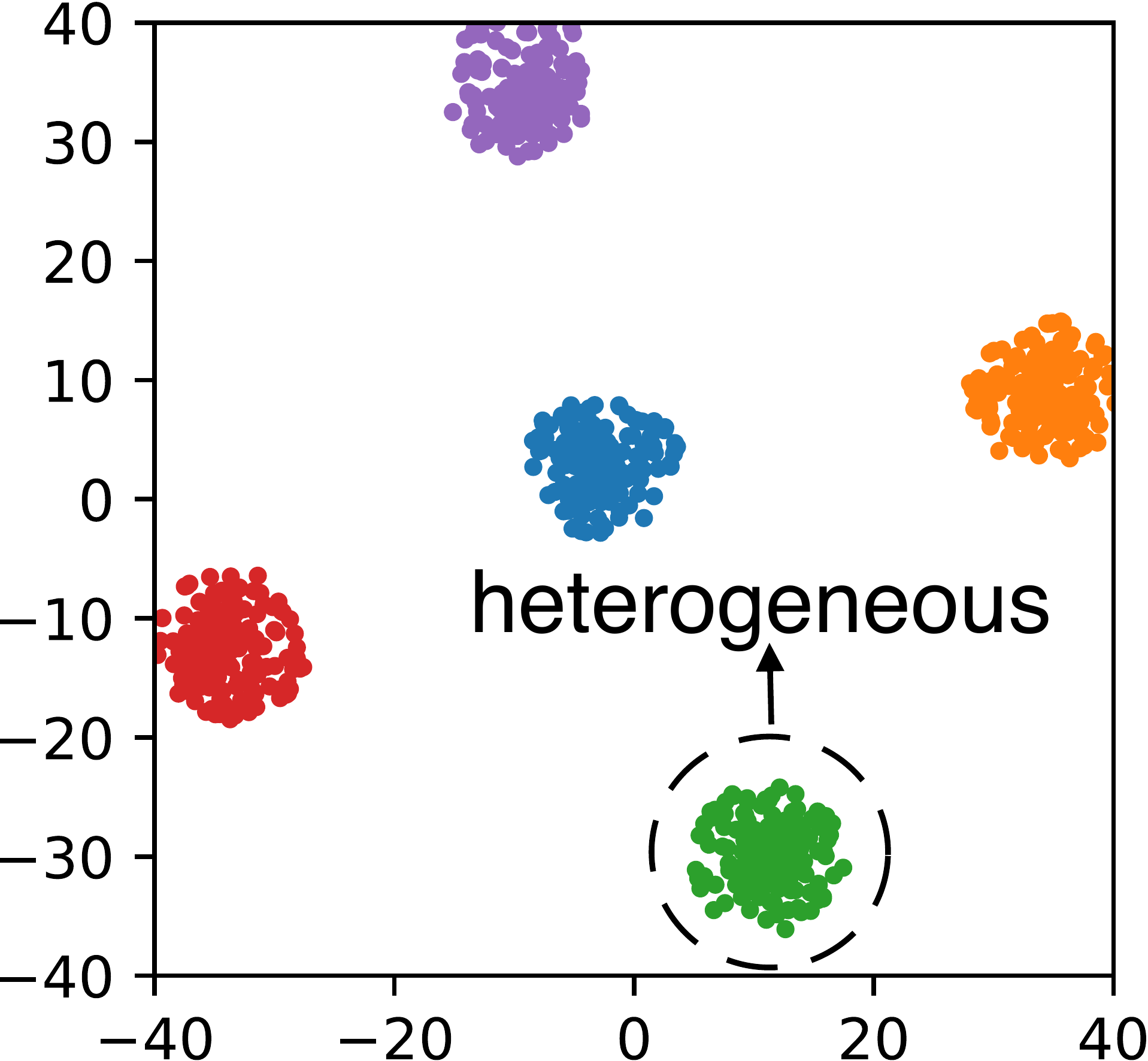}
    \caption{MDC}
    \label{visualization:mdc}
\end{subfigure}
\vspace{-0.5em}
\caption{Feature visualization of ADI and MDC.}
\vspace{-1.2em}
\label{supp:visualization}
\end{figure}

In this section, we provide the comparison between the fake data generated by our MDC and ADI~\cite{yin2020dreaming}. Though both MDC and ADI~\cite{yin2020dreaming} aim to diversify fake images, their manners are quite different: our MDC manipulates the distances among features of fake images, while ADI enlarges disagreement between the student model and the teacher model. 
Fig.\,\ref{supp:visualization} shows feature visualization of ADI and our MDC: the features of MDC scatter a lot while ADI is in a dense concentration. Besides, using 5,120 synthetic images, our MDC obtains 63.77\% top-1 accuracy with ResNet-18 on ImageNet, while ADI only has 54.97\% (we use the official code of ADI). Thus, our MDC can produce more diverse synthetic images as well as better performance.

\section{Other Preprocessing in LOR}

In this section, we report the results of using other preprocessing operations. Tab.\,\ref{supp:Operations} shows no increase from flip and rotation. This is because our LOR aims to capture informative content at different scales and positions of the synthetic images. Flip and rotation may not benefit this goal.
\begin{table}[!ht]
\vspace{-1em}
\centering
\begin{tabular}{c|ccccc}
\hline
Operations & L & L+F   & L+R & L+R+F  \\ \hline
Acc (\%) & 66.47 &  66.34   & 66.31 &  66.42 \\ \hline
\end{tabular}
\vspace{-0.5em}
\caption{Results of 4-bit ResNet-18 on ImageNet when adding flip and rotation to our LOR. ``L'': LOR; ``F'': flip; ``R'': rotation.}
\label{supp:Operations}
\end{table}

\section{Data Amount}

Tab.\,\ref{data-amount} shows the ablation on amount of synthetic images. Note that we achieve 65.87\% accuracy using only 256 images, better than all previous methods, such as the SOTA GZNQ with 64.50\% using 100,000 images.

\begin{table}[ht]
\vspace{-0.5em}
\centering
\begin{tabular}{c|cccccc}
\hline
Amount & 256 & 1,280     & 5,120 &   10,000 & 20,000 \\ \hline
Acc (\%) & 65.87 &  66.14      & 66.47 &  66.49 & 66.50  \\ \hline
\end{tabular}
\vspace{0.5em}
\caption{Performance of our IntraQ \textit{w.r.t.} different amounts of synthetic images (4-bit ResNet-18 on ImageNet).}
\label{data-amount}
\end{table}

\section{More Results}

Tab.\,\ref{supp:comparsion-resnet} shows more comparisons with recent ZSQ methods including Qimera~\cite{choi2021qimera} and SQuant~\cite{guo2021squant}. We report the top-1 accuracy of 4-bit ResNet on ImageNet. Note that SQuant sets the input of the last layer to 8-bit while our IntraQ and Qimera quantize all layers to 4-bit.

\begin{table}[ht]
\centering
\begin{subtable}{\linewidth}\centering
\begin{tabular}{c|c|c|c}
\hline 
Bit-width  & Method  & Generator   & Acc. (\%)\\ \hline \hline
& Real data &  -  & 67.89\\
& Qimera   & \Checkmark & 63.84 \\
& SQuant &  \XSolidBrush  & 66.14\\
\multirow{-4}{*}{W4A4} & \textbf{IntraQ} (Ours)            &  \XSolidBrush   & \textbf{66.47} \\ \hline
\end{tabular}
\captionsetup{font={normalsize}}
\end{subtable}
\vspace{-0.5em}
\caption{Results of ResNet-18 on ImageNet. WBAB indicates the weights and activations are quantized to B-bit.}
\vspace{-1.0em}
\label{supp:comparsion-resnet}
\end{table}

\end{document}